# A Concentration Theorem for Projections

**Sanjoy Dasgupta**
UC San Diego
dasgupta@cs.ucsd.edu

**Daniel Hsu**
UC San Diego
djhsu@cs.ucsd.edu

**Nakul Verma**
UC San Diego
naverma@cs.ucsd.edu

## Abstract

Suppose the random variable $X \in \mathbb{R}^D$ has mean zero and finite second moments. We show that there is a precise sense in which almost all linear projections of $X$ into $\mathbb{R}^d$ (for $d < D$) look like a scale-mixture of spherical Gaussians—specifically, a mixture of distributions $N(0, \sigma^2 I_d)$ where the the $\sigma$ values follow the same distribution as $\|X\|/\sqrt{D}$. The extent of this effect depends upon the ratio of $d$ to $D$, and upon a particular coefficient of eccentricity of $X$'s distribution.

We explore this result in a variety of experiments.

## 1 Introduction

Let $X \in \mathbb{R}^D$ be an arbitrary random variable with mean zero and finite second moments. In this paper, we examine the behavior of "typical" linear projections of $X$ into $\mathbb{R}^d, d < D$.

The first step is to specify a distribution over linear projections from $\mathbb{R}^D$ to $\mathbb{R}^d$. Suppose a $d \times D$ matrix $\Theta$ has entries which are i.i.d. standard normals. It is well-known that with high probability, the rows of this matrix will be approximately orthogonal and have length approximately $\sqrt{D}$; for more details and proof techniques see, for instance, Dasgupta and Gupta (2003). The projection we will use is thus:

$$X \mapsto \frac{1}{\sqrt{D}} \Theta X.$$

An alternative distribution over the projection matrix would be to make its rows the first $d$ basis vectors of a random orthonormal basis of $\mathbb{R}^D$. The distribution we use is quite close to this, and is more convenient to work with analytically and algorithmically.

Let $F_\Theta$ denote the distribution of $X$'s projection, for a particular choice of $\Theta$. We will see that if one were to *average* $F_\Theta$ over over the various possible $\Theta$, the result would be the scale-mixture of spherical Gaussians

$$\overline{F} = \int \nu_\sigma \, \mu(d\sigma) \quad (1)$$

where $\nu_\sigma$ is a shorthand for $N(0, \sigma^2 I_d)$, and $\mu$ follows the distribution of $\|X\|/\sqrt{D}$. Formally, $\mathbb{E}_\Theta[F_\Theta(S)] = \overline{F}(S)$ for any set $S \subseteq \mathbb{R}^d$.

Given the lack of assumptions on $X$, the individual projected distributions $F_\Theta$ could all, for instance, have discrete support. We will show, however, that with high probability over the choice of $\Theta$, the distribution $F_\Theta$ is close to $\overline{F}$ in the following sense: it assigns roughly the same probability mass to every ball $B \subset \mathbb{R}^d$ as does $\overline{F}$. The precise statement is in Theorem 11 but reads approximately like this: for almost all $\Theta$,

$$\sup_{\text{balls } B \text{ in } \mathbb{R}^d} |F_\Theta(B) - \overline{F}(B)| \;\leq\; \widetilde{O}\left(\frac{\mathrm{ecc}(X) d^2}{D}\right)^{1/4}, \quad (2)$$

where $\mathrm{ecc}(X)$ is a specific measure of how eccentric the distribution of $X$ is ($\lambda_{max}/\sigma_\epsilon$ in the theorem statement)[1]. We'll see examples of this value in the next section.

### Implications

Apart from its general insights into data distributions and the enterprise of projection pursuit, we have pursued this result for two rather specific reasons.

The first is curiosity about a widely-observed empirical fact, that a Gaussian distribution is often an accurate density model for one-, two-, or three-dimensional data, but very rarely for high-dimensional data. From the birth weight of babies (Clemens and Pagano, 1999) to the calendar dates of hail and thunder occurrences (Hey and Waylen, 1986), many natural phenomena follow a normal distribution. And yet high-dimensional data is unlikely to be Gaussian, in part because of the high degree of independence this demands (after all, a Gaussian is merely a rotation of a distribution with completely independent coordinates). In a typical application, it might be possible to find a few features that are roughly independent, but as more features are added, the dependencies between them will inevitably grow. See Figure 1 for an illustration of this effect.

[1] The $\widetilde{O}$ and $\widetilde{\Omega}$ notation is used here to suppress factors logarithmic in $1/\epsilon$.

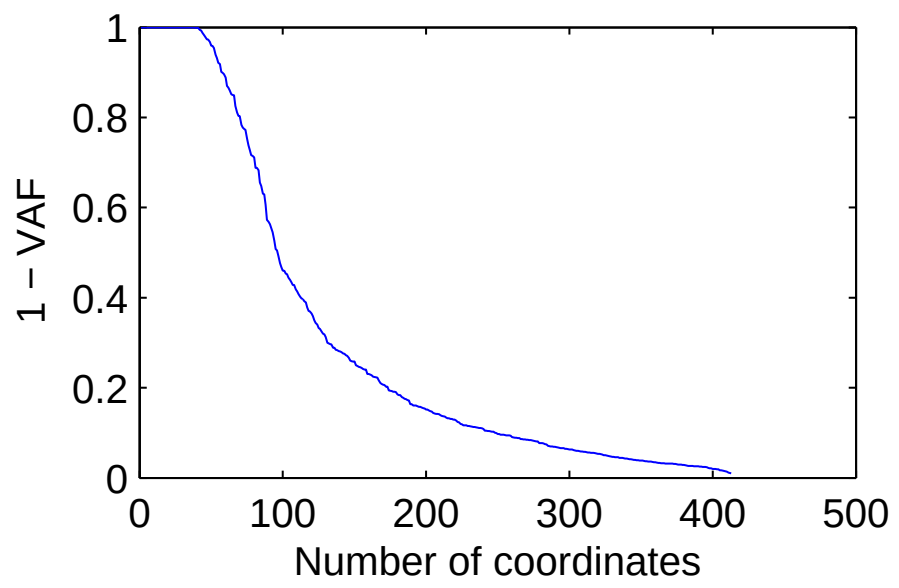

Figure 1: For each coordinate in the MNIST dataset of handwritten "1" digits, this plot shows the fraction of its variance unaccounted for by the best affine combination of the preceding coordinates. The ordering of the coordinates is chosen greedily, by least variance accounted-for.

The result we prove gives a plausible explanation for how high-dimensional distributions that are very far from Gaussian can have low-dimensional projections which are almost Gaussian; and moreover, we quantify the rate at which this effect drops off with increasing $d$.

Our second motivation has to do with the analysis of statistical procedures, and it also explains the particular notion of closeness in distribution. Many learning algorithms do not look too closely at the data but, rather, look only at low-order statistics of the data distribution restricted to simple geometric regions in space. For instance, consider the $k$-means clustering algorithm, whose updates depend only on the zero- and first-order statistics of Voronoi regions determined by the current centers. Its behavior on general data sets is hard to characterize, but its performance on data with Gaussian clusters is much better understood (Dasgupta and Schulman, 2000). Likewise, there has a recent spate of clustering algorithms which are specifically geared towards data whose clusters look approximately Gaussian in terms of their zero-order statistics on balls in space; and which can be rigorously analyzed in this case (Dasgupta, 1999; Arora and Kannan, 2001).

One of the motivations of the present paper is to give a *randomized reduction* from data distributions with fairly general clusters to distributions with better-behaved clusters, and thereby generalize results about the performance of learning algorithms which previously applied only to approximately-Gaussian data. This can be thought of in two ways. Either: the initial process of feature selection can be modeled as being itself a sort of random projection, and thus yielding data whose clusters resemble scale-mixtures of Gaussians in their low-order statistics. Or: random projection can be used as an explicit preprocessing step to specifically produce well-behaved data.

**Previous work**

Our work follows a string of previous results, and draws heavily upon them. The seminal work of Diaconis and Freedman (1984) established this same effect in the case where $X$ has independent coordinates and the projection is to $d = 1$ dimension; in such cases, the projected distribution is close to a single Gaussian (as opposed to a scale-mixture). They also gave an asymptotic result (for $D$ growing to $\infty$) for more general distributions in which most pairs of data points are approximately orthogonal and most data points have approximately unit length.

Sudakov (1978), von Weisäcker (1997), Bobkov (2003), and Naor and Romik (2003) have studied the problem for more general distributions of $X$. These works focus upon $d = 1$ (except for Naor and Romik, who consider general $d$ but define a notion of closeness in distribution which makes the problem essentially one-dimensional), and are based upon various different assumptions on $X$. We closely follow Bobkov's method, and also use ideas from von Weisäcker and Sudakov. Our result is more general than the union of these earlier works in two ways, both of which are crucial for the algorithmic applications mentioned above: (1) we have no constraints, other than finiteness, on the second moments of $X$ (this particular generalization turns out to be easy), and (2) we accommodate the case $d > 1$ (this is the challenging part).

## 2 Examples

Our main result says that most linear projections of $X \in \mathbb{R}^D$ are close to $\overline{F}$, a scale-mixture of Gaussians which is determined only by the distribution of $\|X\|/\sqrt{D}$. We will call this latter distribution the *profile* of $X$.

### 2.1 Discrete distributions

We consider three particular examples: uniform distributions over the vertices of a simplex, a cross-polytope, and a cube in $\mathbb{R}^D$ (Figure 2). In each case, almost all linear projections are near-Gaussian.

**The simplex**

This is the most surprising of the three examples: a discrete distribution in $\mathbb{R}^D$ whose support is of size just $D + 1$, the smallest possible full-dimensional support.

For concreteness, let the vertices be $\{x_0, x_1, \ldots, x_D\}$,

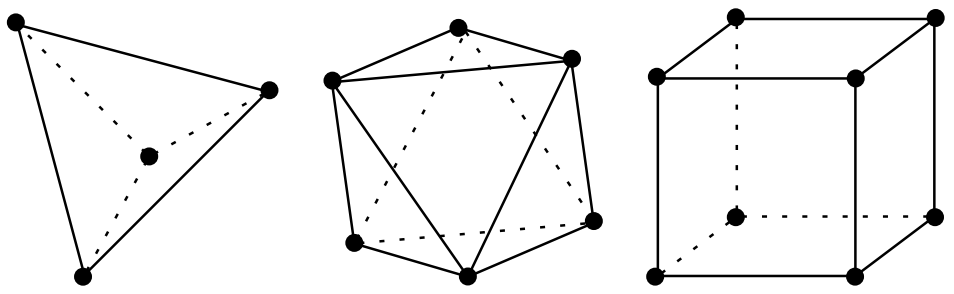

Figure 2: The three-dimensional discrete simplex, cross-polytope, and cube.

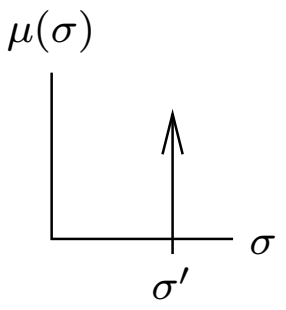

Figure 3: The profile $\mu$ for the uniform distributions over the discrete simplex ($\sigma' = D/(D+1)$), cross-polytope and cube ($\sigma' = 1$).

where

$$x_0 = \frac{1 - \sqrt{D+1}}{\sqrt{D}} \cdot 1_D \ \text{ and }$$

$$x_i = \sqrt{D} e_i \ \text{ for } i = 1, \ldots, D.$$

Here, $1_D$ is the all-ones vector in $\mathbb{R}^D$ and $e_i$ is the $i$th coordinate basis vector.

The crucial fact is that each vertex has the same squared distance $D^2/(D+1)$ to the mean of the distribution and thus the profile $\mu$ puts all of its mass at a single point. This means most linear projections will look Gaussian (rather than a more general scale-mixture).

Specifically, the covariance matrix of the high-dimensional distribution is $(D/(D+1))\, I_D$, and the coefficient of eccentricity is 1. A direct application of Theorem 11 reveals that most projections are close to a single Gaussian, in the sense that the discrepancy on any ball is $\widetilde{O}((d^2/D)^{1/4})$. Figure 4 illustrates this effect.

Notice that the projected distribution has a discrete support of size at most $D+1$. Yet it is almost Gaussian, in the sense that a random sample from this distribution looks just like a random sample from a Gaussian, if you count the number of points in any ball.

In this specific case, we can tighten the bound on the discrepancy. A random projection of the vertices $x_1, x_2, \ldots, x_D$ (ignore $x_0$ for now) is distributed as $D$ independent draws from $N(0, I_d)$: the projection of $x_i$ is

$$\frac{1}{\sqrt{D}} \Theta \left( \sqrt{D} e_i \right) = \Theta_i,$$

the $i$th column of $\Theta$, which has a $N(0, I_d)$ distribution. A standard VC-dimension argument then implies that the fraction of these $D$ projected vertices which fall in any ball is within $O(\sqrt{d(\log D)/D})$ of the probability mass assigned to that ball by $N(0, I_d)$; the crucial technical detail is that the class of balls in $\mathbb{R}^d$ has VC dimension $d+1$. The projection of the remaining vertex $x_0$ can only increase the error by $O(1/D)$.

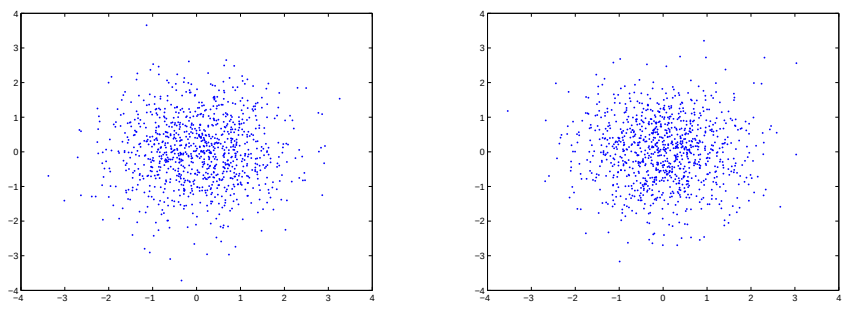

Figure 4: One is the plot of a 2-d projection of the vertices of a 1000-dimensional simplex; the other is the plot of 1001 points sampled from $N(0, I_2)$. Which is which?[1]

**The cross-polytope and cube**

The uniform distributions over the discrete cross-polytope $\{\pm\sqrt{D} e_i : i = 1, \ldots, D\}$ and discrete cube $\{\pm 1\}^D$ are similar: each has covariance $I_D$ and vertices at squared distance $D$ from the center. Again, the profile has mass only at a single point, 1 in these cases. See Figure 3. And again the coefficient of eccentricity is 1, so Theorem 11 shows that most projections are close to a single Gaussian, with the discrepancy on any ball being $O((d^2/D)^{1/4})$.

As with the simplex, a tighter bound for discrepancy can be given in the case of the cross-polytope. We can think of a random projection of the vertices $\sqrt{D} e_i$ as $D$ independent draws from $N(0, I_d)$, call them $\{\theta_1, \ldots, \theta_D\}$; the projections of the remaining $D$ vertices are the negations $\{-\theta_1, \ldots, -\theta_D\}$. With high probability, each half taken separately is close to Gaussian in the sense of being within $\epsilon = O(\sqrt{d(\log D)/D})$ on any ball. So the two halves together are within $2\epsilon$ on any ball.

The uniform distribution over the vertices of the cube $\{-1, +1\}^D$ is different from the previous two examples in that it is a product distribution: its coordinates are independent. Such cases permit special arguments (Diaconis and Freedman, 1984) which show that for 1-d projections, the discrepancy from Gaussian is $O(1/\sqrt{D})$ on any interval of the real line.

## 2.2 Spherically symmetric distributions

Next, we consider the general class of spherically symmetric distributions. This class includes distributions such as the Gaussian, the power-exponential distribution, and Hotelling's T-square distribution. Practitioners in sciences and engineering often prefer this class over the specific case of the Gaussian because it allows for tails that are "heavier" than that of the Gaussian (e.g. Gales and Olson, 1999; Lindsey and Jones, 2000; and see Figure 5).

[1] The second plot shows the Gaussian samples.

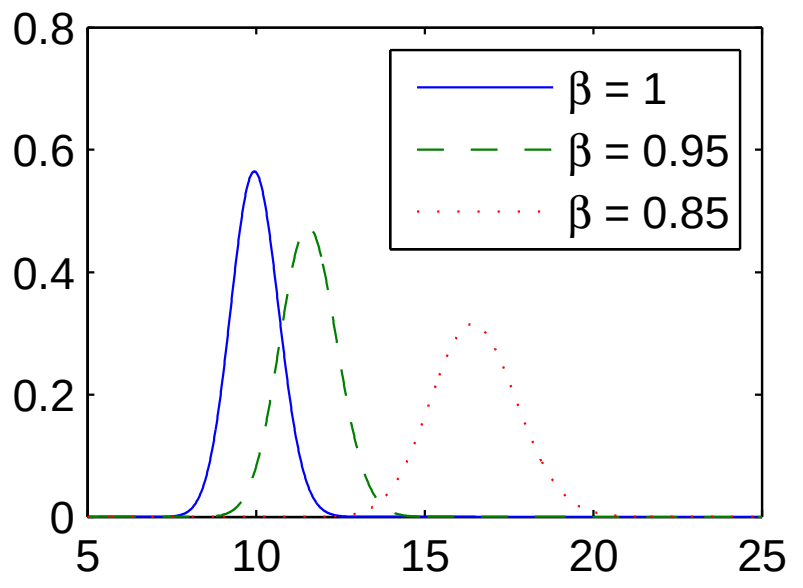

Figure 5: The profile $\mu$ for the power-exponential distributions in $\mathbb{R}^{100}$, parameterized by $\beta$. The Gaussian has $\beta = 1$, while heavy-tailed distributions have $\beta < 1$.

If $X$ has a spherically symmetric distribution centered at the origin, then it can also be written in the form $X = UT$, where $U$ is a random vector uniformly drawn from the $D$-dimensional sphere $S^{D-1}$, and $T$ is a scalar random variable whose distribution is the profile of $X$ (scaled appropriately). We've seen that a random projection will preserve the profile (and therefore the heavy tail) of $X$. This raises an interesting question: can *any* linear projection can alter the tail $T$ of $X$? No, because a linear projection (with orthonormal rows) $\Phi : \mathbb{R}^D \to \mathbb{R}^d$ merely sends

$$X \;\mapsto\; \Phi X \stackrel{d}{=} (\Phi U) T$$

where $\Phi U$ is uniformly distributed over $S^{d-1}$, so the tail $T$ is preserved exactly.

### 2.3 OCR, text, and speech data

Next, we look at low-dimensional projections of three data sets well-known in the machine learning literature: the MNIST database of handwritten digits, the Reuters database of news articles and Mel-frequency cepstral coefficients of the TIMIT data set. Restricting attention to just one cluster from each dataset, we note that projecting the data onto its top principal components suggests the existence of non-Gaussian projections, even though most random projections still look like scale-mixtures of Gaussians (see Figure 6).

### 2.4 Clustered data

Do *all* distributions with finite second moments look like scale-mixtures of Gaussians after a random projection? Far from it! For example, a distribution with two sufficiently-separated clusters has such a high coefficient of eccentricity that the bound on the discrepancy for a particular ball is effectively meaningless (Figure 7). Indeed, in many such cases, the Johnson-Lindenstrauss theorem (1984) dictates that a typical projection will keep the clusters apart, while the result of this paper can more profitably be applied to the individual clusters.

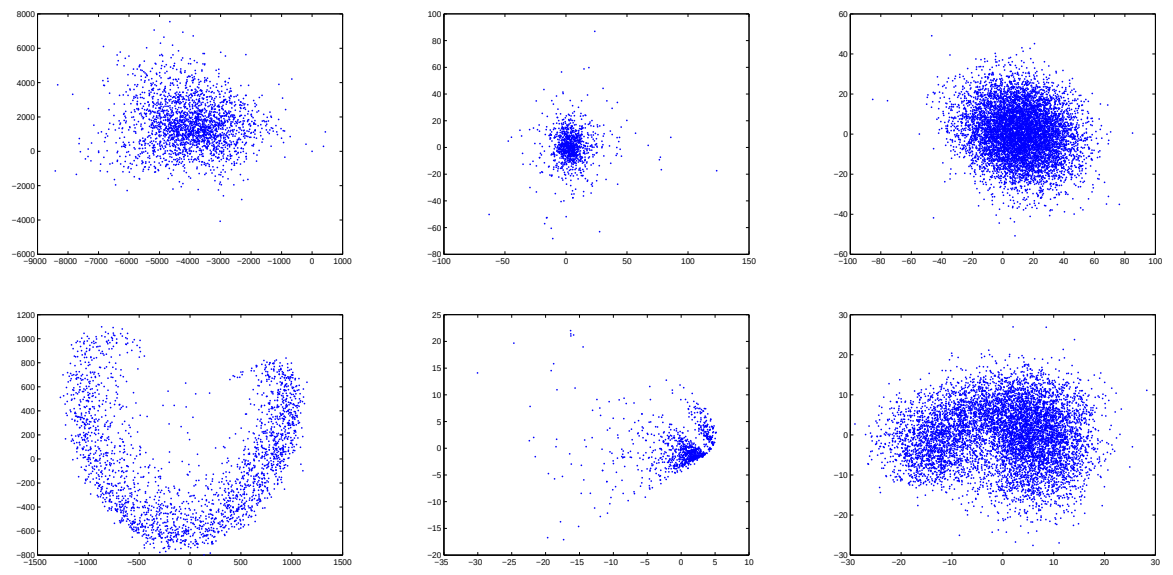

Figure 6: Above: "Typical" two-dimensional projections of handwritten 1's images, word counts of Reuters news articles about Canada, and Mel-frequency cepstral coefficients of the spoken phoneme 's'. Below: The corresponding two-dimensional PCA projections.

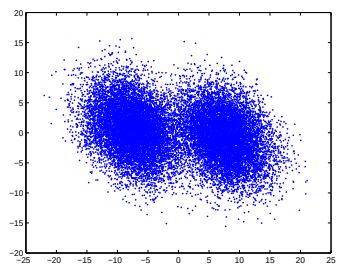

Figure 7: A typical linear projection of a two-cluster (highly eccentric) distribution.

## 3 Proof

The details we omit here (notably the proof of Lemma 7) can be found in the full version of this paper, obtainable from the authors.

### 3.1 Preliminaries

We assume $X \in \mathbb{R}^D$ has mean zero and finite second moments. Let $\mu$ denote the distribution of $\|X\|/\sqrt{D}$; for instance, if $X$ is discrete,

$$\mu(\sigma) \;=\; \mathbb{P}\left[\|X\|^2 = \sigma^2 D\right].$$

Let $\nu_\sigma$ be the $d$-dimensional Gaussian $N(0, \sigma^2 I_d)$, and let $\overline{F}$ be the scale-mixture

$$\overline{F} = \int \nu_\sigma \; \mu(d\sigma).$$

For any $d \times D$ matrix $\Theta$, let $F_\Theta$ denote the distribution of $\frac{1}{\sqrt{D}}\Theta X$. For any open ball $B \subseteq \mathbb{R}^d$, let $F_\Theta(B)$ (which we'll sometimes write $F(\Theta, B)$) be the probability mass that $F_\Theta$ assigns to $B$.

We will consistently use $\|\cdot\|$ to denote Euclidean norm:

$$\|A\|^2 = \begin{cases} \sum_i A_i^2 & \text{if } A \text{ is a vector} \\ \sum_{i,j} A_{ij}^2 & \text{if } A \text{ is a matrix} \end{cases}$$

One last piece of jargon: a function $f : \mathbb{R}^N \to \mathbb{R}$ is $C$-Lipschitz if for all $x, y \in \mathbb{R}^N$,

$$|f(x) - f(y)| \;\le\; C\|x - y\|.$$

### 3.2 Overview

The first part of our proof, following Bobkov (2003), rests crucially upon recent results on concentration of measure, so we start with a brief overview of these.

The familiar Chernoff and Hoeffding bounds say that the *average* of $n$ i.i.d. random variables $X_1, X_2, \ldots, X_n$ is tightly concentrated around its mean, provided the $X_i$ are bounded and $n$ is sufficiently large. But what is so special about the average; what about other functions $f(X_1, \ldots, X_n)$? It turns out that the relevant feature of the average yielding tight concentration is that it is *Lipschitz*.

The following concentration bound applies to *any* Lipschitz function of i.i.d. normal random variables. One good reference for this is Ledoux (2001, page 41, 2.35).

**Theorem 1 (Concentration bound)** *Let $\gamma_N$ denote the distribution $N(0, I_N)$. Suppose the function $f : \mathbb{R}^N \to \mathbb{R}$ is $C$-Lipschitz. Then*

$$\gamma_N\{z : |f(z) - \mathbb{E}[f]| \ge r\} \;\le\; 2e^{-r^2/2C^2}.$$

In our case, the random variable with a $N(0, I_N)$ distribution is the matrix $\Theta$ (so $N = dD$). Here is an outline of our argument.

1. Fix a ball $B \subseteq \mathbb{R}^d$. The first observation is that $\mathbb{E}_\Theta[F_\Theta(B)] = \overline{F}(B)$: *in expectation*, $F_\Theta$ assigns the desired probability mass to $B$.

2. We would like to conclude that $F_\Theta(B)$ is very close to $\overline{F}(B)$ for typical $\Theta$, but this doesn't immediately follow from the concentration bound since $F(\Theta, B)$ may not be Lipschitz (for fixed $B$, variable $\Theta$).

3. So instead, as was done for one-dimensional projections in Bobkov (2003), we introduce a smoothed version of $F_\Theta$. We call it $\widetilde{F}_\Theta$, and we show that *it* is concentrated around its expected value.

4. Then we need to relate $\widetilde{F}_\Theta$ to $F_\Theta$; this is the main technical portion of the proof.

5. So for a fixed ball $B$, for almost all $\Theta$, $F_\Theta(B) \approx \overline{F}(B)$. But we want to show that $F_\Theta(B) \approx \overline{F}(B)$ for *all* balls $B \subseteq \mathbb{R}^d$ simultaneously. To do so, we explicitly construct a finite set of balls $B_1, \ldots, B_M$ with the property that if $F_\Theta$ is close to $\overline{F}$ on these balls, then it is close to $\overline{F}$ on all balls. We finish by taking a union bound over the $B_i$.

### 3.3 The expectation of $F_\Theta$

Let $\Theta$ be a $d \times D$ matrix with i.i.d. $N(0, 1)$ entries. Recall $F_\Theta$ is the distribution of the projected random variable

$$X \;\mapsto\; \frac{1}{\sqrt{D}}\,\Theta X.$$

**Lemma 2** *Fix any $x \in \mathbb{R}^D$. As $\Theta$ varies, the distribution of $\frac{1}{\sqrt{D}}\Theta x$ is $N(0, \frac{\|x\|^2}{D} I_d)$.*

*Proof.* Any linear transformation of a Gaussian is Gaussian, so $\frac{1}{\sqrt{D}}\Theta x$ has a Gaussian distribution. Its mean and second moments are easily checked. ∎

For any ball $B \subseteq \mathbb{R}^d$, define

$$\begin{aligned} F_\Theta(B) = F(\Theta, B) = \mathbb{P}_X\left[\frac{\Theta X}{\sqrt{D}} \in B\right] \\ = \mathbb{E}_X\left[\mathbb{1}\left(\frac{\Theta X}{\sqrt{D}} \in B\right)\right], \end{aligned}$$

where $\mathbb{1}(\cdot)$ denotes the indicator function.

**Lemma 3** *Fix any $B \subseteq \mathbb{R}^d$. As $\Theta$ varies, the expected value of $F_\Theta(B)$ is $\overline{F}(B)$.*

*Proof.* A random draw of $X$ can be achieved in two steps, by first picking $\sigma = \|X\|/\sqrt{D}$ according to $\mu$, and then picking $X$ subject to $\|X\|^2 = \sigma^2 D$.

$$\begin{aligned} &\mathbb{E}_\Theta[F(\Theta, B)] = \mathbb{E}_\Theta\left[\mathbb{E}_X\left[\mathbb{1}\left(\frac{\Theta X}{\sqrt{D}} \in B\right)\right]\right] \\ &= \mathbb{E}_X\left[\mathbb{E}_\Theta\left[\mathbb{1}\left(\frac{\Theta X}{\sqrt{D}} \in B\right)\right]\right] \\ &= \mathbb{E}_\sigma\left[\mathbb{E}_X\left[\mathbb{E}_\Theta\left[\mathbb{1}\left(\frac{\Theta X}{\sqrt{D}} \in B\right)\right] \;\middle|\; \|X\|^2 = \sigma^2 D\right]\right] \\ &= \mathbb{E}_\sigma\left[\mathbb{E}_X\left[\nu_\sigma(B) \mid \|X\|^2 = \sigma^2 D\right]\right] \\ &= \mathbb{E}_\sigma[\nu_\sigma(B)] \;= \overline{F}(B); \end{aligned}$$

the third-last equality is from the previous lemma. ∎

Fix some ball $B$. We can't directly apply the concentration bound to $F(\cdot, B)$ (to show that it is tightly concentrated, over choice of $\Theta$, around its expectation), because this function may not be Lipschitz. The problem is that we allow $X$ to be fairly arbitrary; for instance, it could be uniformly distributed over $k$ support points. Suppose that under projection $\Theta$, exactly one of these support points falls in $B$. Then clearly $F(\Theta, B) = 1/k$. However this projected point might lie right at the boundary of $B$, with the effect that even a tiny perturbation $\Theta \to \Theta'$ causes the point to fall outside $B$, and thus $F(\Theta', B) = 0$. Since $|F(\Theta, B) - F(\Theta', B)|$ cannot be upper-bounded in terms of $\|\Theta - \Theta'\|$, this function is not Lipschitz.

### 3.4 A smoothed version of $F_\Theta$

Fix a ball $B \subseteq \mathbb{R}^d$ and a projection $\Theta$. Consider an experiment in which a point $X$ is randomly drawn and is assigned a score of one if its projection happens to fall in $B$; and a score of zero otherwise. Then $F(\Theta, B) = \mathbb{E}_X\left[\mathbb{1}(\frac{\Theta X}{\sqrt{D}} \in B)\right]$ is the expected score achieved. To get a smoother version of this function, we will assign a fractional score if the projected point doesn't fall exactly in $B$ but is nonetheless close by.

For some value $\Delta > 0$ to be determined, define the function $h_B : \mathbb{R}^d \to [0, 1]$ as follows:

$$h_B(z) = \begin{cases} 1 & \text{if } d(z,B) = 0 \\ 1 - (d(z,B)/\Delta) & \text{if } 0 < d(z,B) \le \Delta \\ 0 & \text{if } d(z,B) > \Delta \end{cases}$$

where $d(z, B) = \inf_{y \in B} \|y - z\|$ is the distance from point $z$ to ball $B$. Clearly $h_B$ is $(1/\Delta)$-Lipschitz.

Now, define the smoothed function $\widetilde{F}(\Theta, B)$ as

$$\widetilde{F}(\Theta, B) \;=\; \mathbb{E}_X\left[h_B\left(\frac{\Theta X}{\sqrt{D}}\right)\right].$$

A one-dimensional version of the following lemma was used by Sudakov (1978).

**Claim 4** *Fix a ball $B \subseteq \mathbb{R}^d$. $\widetilde{F}(\cdot, B)$ is $\sqrt{\lambda_{max}/D\Delta^2}$-Lipschitz, where $\lambda_{max}$ is the largest eigenvalue of the covariance $\mathbb{E}_X\left[XX^T\right]$.*

*Proof.* For any projections $\Theta, \Theta'$,

$$\begin{aligned}
&|\widetilde{F}(\Theta, B) - \widetilde{F}(\Theta', B)| \\
&= \left|\mathbb{E}_X\left[h_B\left(\frac{\Theta X}{\sqrt{D}}\right) - h_B\left(\frac{\Theta' X}{\sqrt{D}}\right)\right]\right| \\
&\le \mathbb{E}_X\left[\left|h_B\left(\frac{\Theta X}{\sqrt{D}}\right) - h_B\left(\frac{\Theta' X}{\sqrt{D}}\right)\right|\right] \\
&\le \frac{1}{\Delta} \cdot \mathbb{E}_X\left[\left\|\frac{\Theta X}{\sqrt{D}} - \frac{\Theta' X}{\sqrt{D}}\right\|\right] \quad (h_B \text{ is } (1/\Delta)\text{-Lipschitz}) \\
&= \frac{1}{\Delta\sqrt{D}} \cdot \mathbb{E}_X\left[\|(\Theta - \Theta')X\|\right] \\
&\le \frac{1}{\Delta\sqrt{D}} \cdot \sqrt{\mathbb{E}_X\left[\|(\Theta - \Theta')X\|^2\right]} \\
&\le \frac{1}{\Delta\sqrt{D}} \cdot \sqrt{\lambda_{max}\|\Theta - \Theta'\|^2} \\
&= \frac{\sqrt{\lambda_{max}}}{\Delta\sqrt{D}} \cdot \|\Theta - \Theta'\|,
\end{aligned}$$

as claimed. ∎

The concentration bound (Theorem 1) gives

**Claim 5** *Fix any ball $B \subset \mathbb{R}^d$, and any $\epsilon > 0$. When $\Theta$ is picked at random,*

$$\mathbb{P}_\Theta\left[|\widetilde{F}(\Theta, B) - \mathbb{E}_\Theta \widetilde{F}(\Theta, B)| \ge \epsilon\right] \;\le\; 2e^{-\epsilon^2\Delta^2 D/2\lambda_{max}}.$$

The problem is that we are interested in the original functions $F_\Theta$ rather than their smoothed counterparts. To relate the two, we use:

$$F_\Theta(B) \;\le\; \widetilde{F}_\Theta(B) \;\le\; F_\Theta(B_\Delta)$$

where $B_\Delta$ is a shorthand for the Minkowski sum $B + B(0, \Delta)$ (to put it simply, grow the radius of $B$ by $\Delta$). By abuse of notation, let $B_{-\Delta}$ be the ball with the same center as $B$ but whose radius is smaller by $\Delta$ (this might be the empty set). Then:

**Corollary 6** *Fix any ball $B \subset \mathbb{R}^d$, and any $\epsilon > 0$. When $\Theta$ is picked at random,*

$$\begin{aligned}
&\mathbb{P}_\Theta\left[\overline{F}(B_{-\Delta}) - \epsilon \le F(\Theta, B) \le \overline{F}(B_\Delta) + \epsilon\right] \\
&\qquad\qquad \ge\; 1 - 2e^{-\epsilon^2\Delta^2 D/2\lambda_{max}}.
\end{aligned}$$

It is necessary, therefore, to relate $\overline{F}(B)$ to $\overline{F}(B_\Delta)$.

### 3.5 Relating the probability mass of $B$ to that of $B_\Delta$

Recall $\overline{F}$ is the scale-mixture

$$\overline{F} \;=\; \int \nu_\sigma \, \mu(d\sigma)$$

where $\nu_\sigma$ is the spherical Gaussian $N(0, \sigma^2 I_d)$. As a first step towards relating $\overline{F}(B)$ and $\overline{F}(B_\Delta)$, we relate $\nu_\sigma(B)$ and $\nu_\sigma(B_\Delta)$.

If $\Delta$ is small enough, then $\nu_\sigma(B_\Delta)$ will not be too much larger than $\nu_\sigma(B)$. But how small exactly does $\Delta$ need to be? There are two effects that come into play.

1. The Gaussian $\nu_\sigma$ looks a lot like a thin spherical shell of radius $\sigma\sqrt{d}$. So it is important to deal properly with balls of radius approximately $\sigma\sqrt{d}$.
2. If $B$ has radius $r$ then $\text{vol}(B_\Delta) \;=\; \text{vol}(B) \cdot \left(\frac{r+\Delta}{r}\right)^d$. So if we want the probability mass of $B_\Delta$ to be at most $(1 + \epsilon)$ times that of $B$, we need $\Delta = O(r\epsilon/d)$.

These two considerations tell us that we need $\Delta \le \epsilon\sigma/\sqrt{d}$. The second also has the troubling consequence that any value of $\Delta$ we choose will only work for balls of radius $> \Delta d$. To get around this, we make sure that $\Delta$ is sufficiently small that any ball of radius less than $\Delta d$ has insignificant (less than $\epsilon$) probability mass.

The following key technical lemma is sketched in the appendix and proved in the full version of the paper.

**Lemma 7** *Pick any* $0 < \epsilon < 1$ *and any* $\sigma > 0$*. If*

$$\Delta \leq \frac{\sigma}{2\sqrt{d}} \cdot \ln\left(1 + \frac{\epsilon}{8}\right) \cdot \frac{1}{1 + \sqrt{\frac{2}{d} \ln \frac{8}{\epsilon}}},$$

*then* $\nu_\sigma(B_\Delta) \leq \nu_\sigma(B) + \epsilon$ *for any ball* $B$*.*

Finally, we consider the scale-mixture rather than just individual components $\nu_\sigma$.

**Corollary 8** *Pick any* $0 < \epsilon < 1$ *and a threshold* $\sigma_\epsilon > 0$ *such that* $\mu\{\sigma : \sigma < \sigma_\epsilon\} \leq \epsilon$*. If*

$$\Delta \leq \frac{\sigma_\epsilon}{2\sqrt{d}} \cdot \ln\left(1 + \frac{\epsilon}{8}\right) \cdot \frac{1}{1 + \sqrt{\frac{2}{d} \ln \frac{8}{\epsilon}}},$$

*then* $\overline{F}(B_\Delta) \leq \overline{F}(B) + 2\epsilon$*.*

*Proof.* We can rewrite $\overline{F}$ as $\mathbb{E}_\sigma[\nu_\sigma]$, where the expectation is taken over $\sigma$ drawn according to $\mu$.

$$\begin{aligned} &\overline{F}(B_\Delta) - \overline{F}(B) \\ &\quad = \mathbb{E}_\sigma[\nu_\sigma(B_\Delta)] - \mathbb{E}_\sigma[\nu_\sigma(B)] \\ &\quad \leq \mathbb{E}_\sigma[\nu_\sigma(B_\Delta) - \nu_\sigma(B) \mid \sigma \geq \sigma_\epsilon] + \mathbb{P}_\sigma(\sigma < \sigma_\epsilon) \\ &\quad \leq 2\epsilon, \end{aligned}$$

as claimed. ∎

At this stage, we have shown that for any given $B$, almost all projections $\Theta$ have $F_\Theta(B) \approx \overline{F}(B)$.

**Theorem 9** *Pick any* $0 < \epsilon < 1$ *and* $\sigma_\epsilon > 0$ *such that* $\mu\{\sigma : \sigma < \sigma_\epsilon\} \leq \epsilon$*. Pick any ball* $B \subseteq \mathbb{R}^d$*. Then*

$$\mathbb{P}_\Theta\left[|F_\Theta(B) - \overline{F}(B)| > \epsilon\right] \leq \exp\left\{-\widetilde{\Omega}\left(\frac{\epsilon^4 D}{d} \cdot \frac{\sigma_\epsilon^2}{\lambda_{max}}\right)\right\}$$

It remains to prove this for all balls *simultaneously*.

### 3.6 Uniform convergence for all balls

We follow a standard method for proving uniform convergence: we carefully choose a small finite set of balls $B_0, \ldots, B_M \subseteq \mathbb{R}^d$ such that if the concentration property (Theorem 9) holds on these $B_i$'s, then it holds for *all* balls in $\mathbb{R}^d$. Specifically, our $B_i$'s have the following property:

> For any ball $B \subseteq \mathbb{R}^d$ there exist $B_i, B_j$ such that $B_i \subseteq B \subseteq B_j$ and $\overline{F}(B_j) - \overline{F}(B_i) \leq \epsilon$.

It follows that if $F_\Theta(B_i) \approx \overline{F}(B_i)$ for the finite set of balls $B_i$, then $F_\Theta(B) \approx \overline{F}(B)$ for all balls $B \subseteq \mathbb{R}^d$. (Actually, there is a slight complication in that we have to separately deal with balls centered very far from the origin.)

Here's the construction of the balls $B_1, \ldots, B_M$, for some parameters $c, \epsilon_o$ to be determined:

1. Place a grid with resolution (spacing) $2\epsilon_o$ on $[-c\sqrt{d}, c\sqrt{d}]^d$.
2. At each point on the grid, create a set of balls centered at that point, with radii $\epsilon_o\sqrt{d}, 2\epsilon_o\sqrt{d}, \ldots, (2c + 2\epsilon_o)\sqrt{d}$.

The total number of balls is then $M = \left(\frac{c\sqrt{d}}{\epsilon_o}\right)^d \cdot \frac{2c+2\epsilon_o}{\epsilon_o}$. For good measure, add one final ball: $B_0 = \mathbb{R}^d$.

**Lemma 10** *Set* $c \geq \sqrt{\lambda_{avg}/2\epsilon}$ *and* $\epsilon_o \leq \Delta/(4\sqrt{d})$*. Pick any ball* $B \subseteq \mathbb{R}^d$ *centered in* $B(0, c\sqrt{d})$*. Then there exist* $B_i, B_j$ *such that*

$$B_i \subseteq B \subseteq B_j$$

*and* $\overline{F}(B_j) - \overline{F}(B_i) \leq 2\epsilon$*.*

*Proof.* Say $B = B(x, r)$, with $x \in B(0, c\sqrt{d})$.

Case 1: $r \leq 2c\sqrt{d}$.

By construction, there is a grid point $\tilde{x}$ which differs from $x$ by at most $\epsilon_o$ on each coordinate, so $\|x - \tilde{x}\| \leq \epsilon_o\sqrt{d}$. Let $B^{in} = B(\tilde{x}, r_1)$ be the largest of the $B_i$'s centered at $\tilde{x}$ and *contained inside* $B$. Likewise let $B^{out} = B(\tilde{x}, r_2)$ be the smallest of the $B_i$'s centered at $\tilde{x}$ and *containing* $B$. Again by construction,

$$\begin{aligned} r_1 &\geq r - \|x - \tilde{x}\| - \epsilon_o\sqrt{d} \geq r - 2\epsilon_o\sqrt{d} \\ r_2 &\leq r + \|x - \tilde{x}\| + \epsilon_o\sqrt{d} \leq r + 2\epsilon_o\sqrt{d} \end{aligned}$$

Since $\epsilon_o \leq \Delta/(4\sqrt{d})$, we have $B^{out} \subseteq B^{in}_\Delta$ and thus, by corollary 8, $\overline{F}(B^{out}) - \overline{F}(B^{in}) \leq 2\epsilon$.

Case 2: $r > 2c\sqrt{d}$.

In this case, $B$ is contained in $B_0 = \mathbb{R}^d$ and contains the ball $B_i$ which is centered at the origin and has radius $c\sqrt{d}$. It remains to show that $\overline{F}(B_i) \geq 1 - 2\epsilon$.

Let $Z$ be a random draw from $\overline{F} = \int \nu_\sigma \mu(d\sigma)$.

$$\begin{aligned} \mathbb{E}\left[\|Z\|^2\right] &= \int \sigma^2 d\, \mu(d\sigma) \\ &= \frac{d}{D}\mathbb{E}\left[\|X\|^2\right] = d\lambda_{avg}, \end{aligned}$$

where $\lambda_{avg}$ is the average eigenvalue of $\mathbb{E}[XX^T]$. Since $c^2 \geq \lambda_{avg}/(2\epsilon)$, by Markov's inequality

$$\mathbb{P}\left[\|Z\|^2 \geq c\sqrt{d}\right] \leq \frac{\mathbb{E}\left[\|Z\|^2\right]}{c^2 d} \leq 2\epsilon$$

and so $\overline{F}(B_i) = 1 - \mathbb{P}[\|Z\|^2 \geq c\sqrt{d}] \geq 1 - 2\epsilon$. ∎

$B(0, c\sqrt{d})$ is so large that less than an $\epsilon$ fraction of $\overline{F}$ lies outside it. We need a separate argument to deal with balls which are centered outside $B(0, c\sqrt{d})$, and we finally get

**Theorem 11** *Suppose $X$ has mean zero and finite second moments. Define $F_\Theta, \mu, \overline{F}$ as above. Pick any $0 < \epsilon < 1$ and $\sigma_\epsilon > 0$ such that $\mu\{\sigma : \sigma < \sigma_\epsilon\} \leq \epsilon$. Then*

$$\mathbb{P}_\Theta \left[ \sup_{\text{balls } B \subseteq \mathbb{R}^d} |F_\Theta(B) - \overline{F}(B)| > \epsilon \right] \leq \left( \widetilde{O} \left( \frac{d^3}{\epsilon^3} \frac{\lambda_{avg}}{\sigma_\epsilon^2} \right) \right)^{d/2} \exp \left\{ -\widetilde{\Omega} \left( \frac{\epsilon^4 D}{d} \cdot \frac{\sigma_\epsilon^2}{\lambda_{max}} \right) \right\}$$

*where $\lambda_{avg}$ and $\lambda_{max}$ are the average and maximum eigenvalues of the covariance $\mathbb{E}[XX^T]$.*

**Acknowledgements**

We are grateful to Yoav Freund for suggesting this problem, to the anonymous reviewers for their extremely helpful feedback, and to the NSF for grant IIS-0347646.

# 4 Appendix: proof sketch for Lemma 7

Let $B = B(x, r)$, so that $B_\Delta = B(x, r + \Delta)$. For some $\sigma > 0$, we will compare $\nu_\sigma(B)$ with $\nu_\sigma(B_\Delta)$. Note that

$$\nu_\sigma(B_\Delta) = \frac{1}{(2\pi)^{d/2}\sigma^d} \int_{B(0,r+\Delta)} e^{-\|x+z\|^2/2\sigma^2} dz$$

can be rewritten as

$$\frac{1}{(2\pi)^{d/2}\sigma^d} \left( \frac{r+\Delta}{r} \right)^d \int_{B(0,r)} e^{-\|x+y+(\Delta/r)y\|^2/2\sigma^2} dy$$

under the change of variable $y = z \cdot \frac{r}{r+\Delta}$. To relate this integral to $\nu_\sigma(B)$, we divide it into two parts: $\{y : \|x + y\| \geq c\sigma\sqrt{d}\}$ (faraway points) and $\{y : \|x + y\| < c\sigma\sqrt{d}\}$ (nearby points). Here $c \approx 2$ is a constant to be determined.

**Lemma 12** *Pick any $0 < \epsilon_1 < 1/2$, and set*

$$\Delta \leq \frac{\sigma}{2\sqrt{d}} \cdot \ln(1 + \epsilon_1) \cdot \frac{1}{1 + \sqrt{\frac{2}{d} \ln \frac{1}{\epsilon_1}}}.$$

*Then, for an appropriate choice of the constant $c$:*

1. *Faraway points:*
$$\frac{1}{(2\pi)^{d/2}\sigma^d} \int_{\substack{y \in B(0,r) \\ \|x+y\| \geq c\sigma\sqrt{d}}} e^{-\|x+y+(\Delta/r)y\|^2/2\sigma^2} dy$$
*is at most $(1 + \epsilon_1)\epsilon_1$.*

2. *Nearby points:*
$$\frac{1}{(2\pi)^{d/2}\sigma^d} \int_{\substack{y \in B(0,r) \\ \|x+y\| < c\sigma\sqrt{d}}} e^{-\|x+y+(\Delta/r)y\|^2/2\sigma^2} dy$$
*is at most $(1 + \epsilon_1)\nu_\sigma(B)$.*

3. *Combining these and choosing $\epsilon_1 = \epsilon/8$,*
$$\nu_\sigma(B_\Delta) \leq \left( \frac{r+\Delta}{r} \right)^d \left( \nu_\sigma(B) + \frac{3\epsilon}{8} \right).$$

This bound is reasonable if $r$ is large. On the other hand, if $r$ is tiny, then $\nu_\sigma(B_\Delta)$ is less than $\epsilon$, and the point is moot anyway. The nontrivial case is when $r$ is of intermediate size (details in full paper).